\documentclass[final,3p,times,twocolumn]{elsarticle}

\usepackage{epsfig}
\usepackage{hyperref}   
\usepackage{url} 
\usepackage{graphicx}%
\usepackage{multirow}%
\usepackage{amsmath,amssymb,amsfonts}%
\usepackage{amsthm}%
\usepackage{mathrsfs}%
\usepackage[title]{appendix}%
\usepackage{textcomp}%
\usepackage{manyfoot}%
\usepackage{booktabs}%
\usepackage{algorithmicx}%
\usepackage{algpseudocode}%
\usepackage{listings}%
\usepackage[ruled,vlined]{algorithm2e}

\usepackage{booktabs}
\usepackage{multirow}
\usepackage[table,xcdraw]{xcolor}
\usepackage{times,epsf,mathtools,xcolor,amssymb,amsmath,hyperref, enumitem,wrapfig, subcaption, comment, cleveref, appendix, float}
\usepackage{tabu}
\usepackage{xcolor}
\usepackage{graphicx}
\usepackage{tikz}
\usetikzlibrary{shapes.geometric, arrows}

\definecolor{LightCyan}{rgb}{0.88,1,1}

\begin{document}

\begin{frontmatter}

\title{HOG-CNN: Integrating Histogram of Oriented Gradients with Convolutional Neural Networks for Retinal Image Classification}

\author[aff1]{Faisal Ahmed\corref{cor1}}
\ead{ahmedf9@erau.edu}  



 \cortext[cor1]{Corresponding author}

\address[aff1]{Department of Data Science and Mathematics, Embry-Riddle Aeronautical University, 3700 Willow Creek Rd, Prescott, Arizona 86301, USA}

\begin{abstract}
The analysis of fundus images is critical for the early detection and diagnosis of retinal diseases such as Diabetic Retinopathy (DR), Glaucoma, and Age-related Macular Degeneration (AMD). Traditional diagnostic workflows, however, often depend on manual interpretation and are both time- and resource-intensive. To address these limitations, we propose an automated and interpretable clinical decision support framework based on a hybrid feature extraction model called \textit{HOG-CNN}.

Our key contribution lies in the integration of handcrafted Histogram of Oriented Gradients (HOG) features with deep convolutional neural network (CNN) representations. This fusion enables our model to capture both local texture patterns and high-level semantic features from retinal fundus images. We evaluated our model on three public benchmark datasets: APTOS 2019 (for binary and multiclass DR classification), ORIGA (for Glaucoma detection), and IC-AMD (for AMD diagnosis); HOG-CNN demonstrates consistently high performance. It achieves \textbf{98.5\% accuracy and 99.2 AUC} for binary DR classification, and \textbf{94.2 AUC} for five-class DR classification. On the IC-AMD dataset, it attains \textbf{92.8\% accuracy, 94.8\% precision and 94.5 AUC}, outperforming several state-of-the-art models. For Glaucoma detection on ORIGA, our model achieves \textbf{83.9\% accuracy and 87.2 AUC}, showing competitive performance despite dataset limitations.

We show, through comprehensive appendix studies, the complementary strength of combining HOG and CNN features. The model's lightweight and interpretable design makes it particularly suitable for deployment in resource-constrained clinical environments. These results position HOG-CNN as a robust and scalable tool for automated retinal disease screening.

\end{abstract}


\begin{highlights}
\item We propose a novel HOG-CNN framework that integrates Histogram of Oriented Gradients (HOG) with deep CNN features for retinal disease diagnosis.
\item Our model demonstrates state-of-the-art performance in disease classification tasks, achieving up to 98.5\% accuracy and an AUC of 99.2 for diabetic retinopathy (DR) classification on the APTOS dataset, and up to 92.8\% accuracy, precision 94.8\%, and an AUC of 94.5 for age-related macular degeneration (AMD) classification on the IC-Dataset.
\item HOG-CNN demonstrates strong generalization across three retinal diseases—\\AMD, DR, and Glaucoma—using multiple public benchmark datasets.
\item The hybrid design effectively captures both local texture patterns and high-level semantic information from fundus images.
\item Our model offers a lightweight and interpretable alternative to complex deep learning systems, making it suitable for real-world, resource-constrained clinical environments.
\end{highlights}


\end{frontmatter}



\section{Introduction}
\label{sec:introduction}
As of August 2023, the World Health Organization (WHO) reports that over 2.2 billion people globally experience near or distance vision impairment, with at least 1 billion cases being preventable or yet to be addressed. Among the leading causes of vision impairment and blindness are glaucoma, diabetic retinopathy (DR), and age-related macular degeneration (AMD). Specifically, the WHO estimates that glaucoma affects 7.7 million, DR affects 3.9 million, and AMD affects 8 million individuals worldwide. \cite{WHO2023vision}. These figures underscore the significant global burden of serious eye diseases, highlighting the need for improved access to eye care services worldwide. As most patients with eye diseases are not aware of the aggravation of these conditions, early screening and treatment of eye diseases are quite important. Currently, detecting these conditions is a time-consuming and manual process that requires a trained clinician to examine and evaluate digital color fundus images of the retina, which can result in delayed treatment. Therefore, the need for clinical decision-support methods has long been recognized.
Recent advancements have further enhanced diagnostic capabilities, particularly in predicting disease progression. For example, an automated ML model has been shown to effectively predict short-term DR progression using ultra-widefield retinal images, with promising implications for early screening and improved vision outcomes~\cite{silva2024automated}. Additionally, deep learning algorithms have demonstrated strong performance in forecasting the progression to geographic atrophy in AMD patients by analyzing spectral-domain optical coherence tomography images. Integrating clinical data with optical coherence tomography angiography has also led to improvements in DR classification accuracy~\cite{dow2023deep}. Despite these advancements, current ML methods often lack computational efficiency when applied to large datasets and remain limited in interpretability, which is crucial for providing actionable insights to ophthalmologists. To address these limitations, recent approaches such as HOG-CNN leverage histograms of oriented gradients  to enhance both the performance and interpretability of retinal image analysis models.

Driven by this motivation, the past decade has seen extensive use of machine learning (ML) techniques in the analysis of retinal images \cite{li2021applications,ting2019deep}. Significant advancements have been achieved through approaches such as image classification and pattern recognition~\cite{sarhan2020machine}. The emergence and success of convolutional neural networks (CNNs) in image classification further demonstrated the effectiveness of ML in retinal image processing~\cite{ting2019artificial}. Despite these advances, many of these methods fall short in terms of computational efficiency for handling large-scale datasets and lack the interpretability needed to support ophthalmologists in diagnosing diseases.

In this study, we propose a novel hybrid approach for retinal disease classification by incorporating Histogram of Oriented Gradients (HOG) features into deep learning models. Inspired by the foundational work of Dalal and Triggs \cite{dalal2005histograms}, where HOG descriptors were effectively utilized for human detection, we extend this technique to the domain of retinal image analysis. Specifically, we extract HOG features from each retinal fundus image, resulting in a high-dimensional descriptor of size $26{,}244$ per image. These handcrafted features are then fused with deep features obtained from a pre-trained convolutional neural network (CNN), forming a hybrid HOG-CNN architecture. The integration of structural gradient information with learned visual representations enhances the model's ability to discriminate between disease categories. Experimental results of our model HOG-CNN demonstrate that the proposed HOG-CNN model achieves superior performance compared to conventional CNN-based classifiers, highlighting the benefits of combining traditional feature engineering with modern deep learning techniques.

\medskip

\noindent \textbf{Our contributions.}

\begin{itemize}
    \item We propose a novel HOG-CNN framework that integrates Histogram of Oriented Gradients (HOG) with deep CNN features for retinal disease diagnosis.
\item Our model demonstrates state-of-the-art performance in disease classification tasks, achieving up to 98.5\% accuracy and an AUC of 99.2 for diabetic retinopathy (DR) classification on the APTOS dataset, and up to 92.8\% accuracy, precision 94.8\% and an AUC of 94.5 for age-related macular degeneration (AMD) classification on the IC-Dataset.
\item HOG-CNN demonstrates strong generalization across three retinal diseases—AMD, DR, and Glaucoma—using multiple public benchmark datasets.
\item The hybrid design effectively captures both local texture patterns and high-level semantic information from fundus images.
\item Our model offers a lightweight and interpretable alternative to complex deep learning systems, making it suitable for real-world, resource-constrained clinical environments.
\end{itemize}

\begin{figure*}[t!] 
	\centering
	\subfloat[\scriptsize Color retinal fundus image \label{fig:Color-HOG img}]{%
		\includegraphics[width=0.32\linewidth]{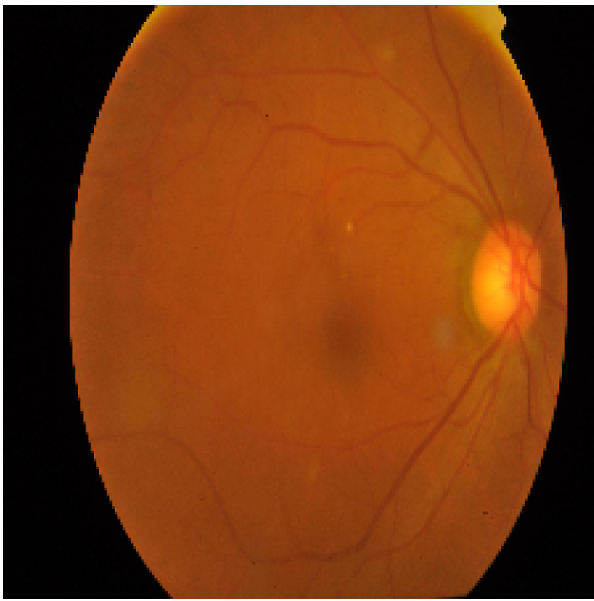}}
	\hfill
	\subfloat[\scriptsize Grayscale retinal fundus image\label{fig:color img}]{%
		\includegraphics[width=0.32\linewidth]{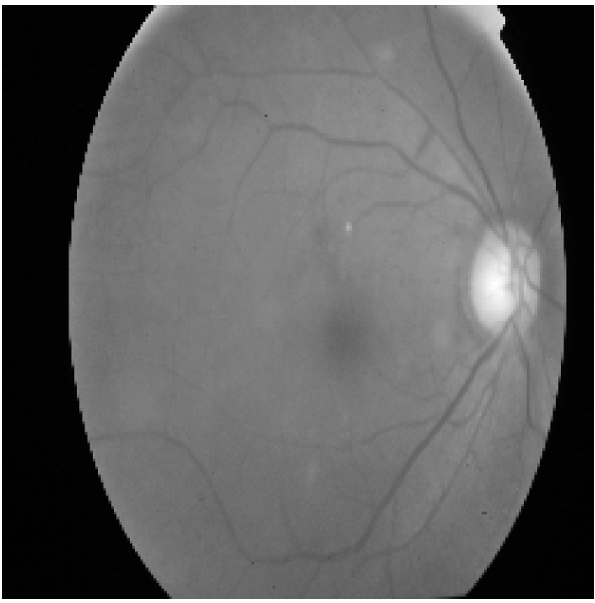}}
	\hfill
	\subfloat[\scriptsize HOG visualization \label{fig:HOG img}]{%
		\includegraphics[width=0.32\linewidth]{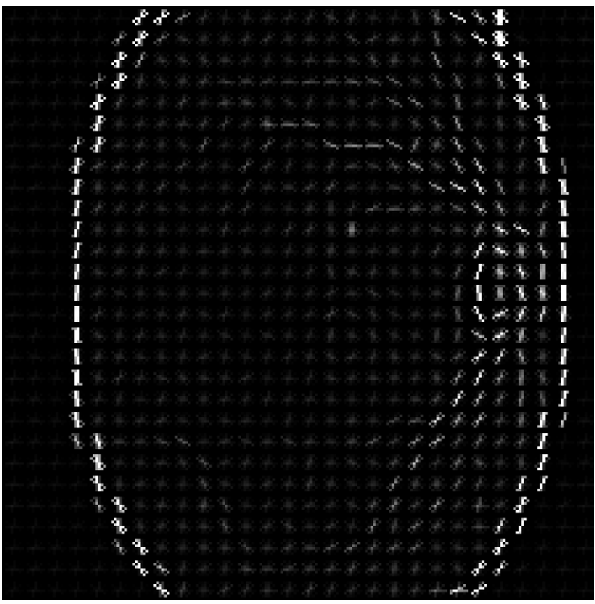}}
	\caption{\footnotesize Illustration of the preprocessing pipeline applied to a fundus image. From left to right: (a) Original color image, (b) grayscale conversion used for HOG computation, and (c) corresponding Histogram of Oriented Gradients (HOG) visualization highlighting edge and texture features.}
	\label{fig:HOG-visualization}
\end{figure*}

\section{Related Work}
\label{sec:related} 

In the past decade, machine learning (ML) tools have been extensively employed in medical image analysis~\cite{singh2020explainable,fourcade2019deep}. In particular, within the domain of retinal image analysis, ML methodologies have demonstrated significant effectiveness~\cite{ting2019artificial}. There are two primary applications of ML in retinal imaging: (1) the diagnosis and grading of diseases, which is typically framed as a classification problem for image data~\cite{pratt2016convolutional}, and (2) lesion detection and segmentation~\cite{li2021applications}. In this paper, we focus exclusively on the diagnosis and grading of retinal diseases by leveraging and integrating Histogram of Oriented Gradients (HOG) features with pre-trained convolutional neural network (CNN) models.

Topological Data Analysis (TDA) is an emerging approach that is increasingly being utilized in medical image analysis, including applications in retinal imaging~\cite{ahmed2025topo, ahmed2023tofi, ahmed2023topological}. Over the past decade, TDA has shown great promise across diverse fields such as image analysis, neurology, cardiology, hepatology, gene-level and single-cell transcriptomics, drug discovery, evolutionary biology, and protein structure analysis. Its strength lies in revealing latent patterns within data, offering new avenues for tasks like image segmentation, object recognition, registration, and reconstruction. In medical imaging, persistent homology (PH)—a key method in TDA—has proven effective in analyzing histopathology slides~\cite{qaiser2019fast,lawson2019persistent,10385822}, fibrin network structures~\cite{berry2020functional}, tumor classification~\cite{crawford2020predicting, yadav2023histopathological}, chest X-ray screening~\cite{ahmed2023topo},  neuronal morphology studies~\cite{kanari2018topological}, brain artery mapping~\cite{bendich2016persistent}, fMRI analysis~\cite{rieck2020uncovering,stolz2021topological}, and genomic inference~\cite{camara2016inference}.

Following the success of CNNs in image classification tasks, deep learning methods have proven to be highly effective in retinal image analysis~\cite{ting2019deep,orlando2018ensemble}. A substantial body of literature exists on deep learning applications in ophthalmology, with comprehensive reviews available in survey papers ~\cite{you2022application,srivastava2023artificial,sarhan2020machine,li2021applications}. Moreover, recent studies have explored advanced ML models for predicting disease progression~\cite{silva2024automated}, and have integrated clinical data with imaging modalities such as OCT-A to improve diabetic retinopathy (DR) classification~\cite{dow2023deep}. Additional reviews highlighting the current state of deep learning in ophthalmology can be found in the following works~\cite{koseoglu2023predictive,ashtari2024glaucoma,nguyen2025deep}.

\section{Methodology} \label{sec:HOG}

In this study, we employ the Histogram of Oriented Gradients (HOG) technique as an effective feature extraction method for the classification of retinal images. The methodology consists of two primary stages: (1) extraction of HOG features from retinal images, and (2) integration of these features with the weights of pre-trained convolutional neural network (CNN) models to form a hybrid deep learning framework for classification. First, HOG descriptors are computed for each retinal image, capturing edge and gradient structure that are characteristic of various retinal conditions. These descriptors are then used to augment the feature representations learned by the CNN models. By combining handcrafted features with deep features, we aim to enhance the discriminative power of the model. In the subsequent subsections, we detail the process of HOG feature extraction and the integration strategy used to combine these features with the pre-trained CNNs.

\begin{figure*}[t!]
    \centering
     \includegraphics[width=.8\linewidth]{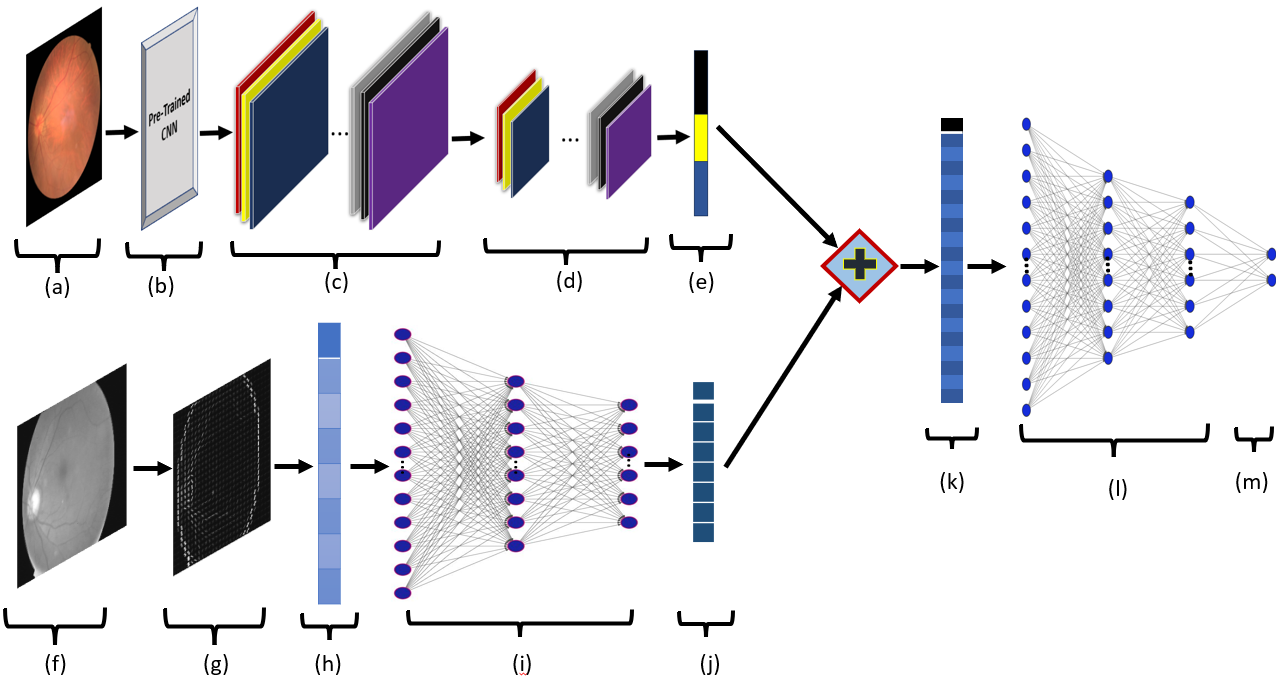}
     \caption{\small {\bf Overview of the proposed model HOG-CNN architecture:} The model accepts two inputs: (a) a color fundus image and (f) a grayscale fundus image. The top pathway processes (a) through (b) a pre-trained CNN, (c) a 64-filter convolutional layer, (d) a $2 \times 2$ max-pooling layer, and (e) a flattening layer. In the bottom pathway, (g) the HOG visualization is extracted from (f), and (h) its feature vector is passed through (i) a feed-forward neural network with three hidden layers of sizes 800, 256, and 128. Outputs from both pathways, (e) and (j), are concatenated to form (k), (l) which is input to a second feed-forward network with hidden layers of sizes 256 and 128, along with a dropout layer (dropout rate = 0.2). The final output layer (m) has size 2 for binary or size 5 for multi-class classification.}
     
     \label{fig:HOG-CNN}
\vspace{-.in}
 \end{figure*}

\subsection{HOG Feature Extraction from Retinal Images}

In this study, we employed the Histogram of Oriented Gradients (HOG) as a feature extraction technique to capture the texture and structural characteristics of retinal fundus images. HOG is a widely-used descriptor for object detection and image classification due to its ability to encode gradient orientation distributions, which are particularly useful in representing local shape and edge information.

The feature extraction pipeline began with data acquisition, denoted as $\mathcal{I} = \{I_1, I_2, \ldots, I_n\}$, where each $I_k$ represents a color fundus image. Each image was resized to a uniform spatial resolution of $224 \times 224$ pixels to ensure consistency across the dataset and to facilitate compatibility with pre-trained convolutional neural network (CNN) models used later in our hybrid framework. This resizing can be denoted as:
\[
I_k' = \text{Resize}(I_k, 224 \times 224)
\]

Following resizing, each RGB image $I_k'$ was converted to a grayscale image $G_k$ using a luminance-preserving transformation:
\[
G_k(x, y) = 0.299 \cdot R(x, y) + 0.587 \cdot G(x, y) + 0.114 \cdot B(x, y)
\]
This grayscale conversion is essential since HOG operates on intensity gradients, and color information does not significantly enhance gradient-based descriptors.

Once converted, the HOG descriptor was computed on each grayscale image. This involved calculating the gradient magnitude and orientation at each pixel using discrete derivative masks. The image was then partitioned into small spatial regions known as \emph{cells}, each of size $8 \times 8$ pixels. Within each cell, a histogram of gradient directions was computed using 9 orientation bins (covering $0^\circ$ to $180^\circ$). These histograms describe the distribution of edge directions within the local area.

To account for changes in illumination and contrast, these local histograms were grouped into overlapping \emph{blocks} of $2 \times 2$ cells. Each block's histograms were concatenated and normalized using the L2-Hys norm to produce a robust local feature vector. Mathematically, this can be expressed as:
\[
F_k = \text{HOG}(G_k; o=9, p=8 \times 8, b=2 \times 2)
\]
where $F_k \in \mathbb{R}^d$ is the resulting high-dimensional HOG descriptor for image $k$, and $d$ denotes the length of the flattened feature vector. The value of $d$ depends on the total number of cells and blocks given the fixed image resolution. In our experiment, $d=26{,}244$, so $F_k \in \mathbb{R}^{26{,}244}$; that is, from each grayscale image, we extract a $26{,}244-$dimensional vector. 

This process was repeated for all $n$ images in the dataset, resulting in a feature matrix:
\[
\mathbf{F} = 
\begin{bmatrix}
F_1 \\
F_2 \\
\vdots \\
F_n
\end{bmatrix}
\in \mathbb{R}^{n \times d}
\]

The matrix $\mathbf{F}$ was then stored in tabular format using a DataFrame structure and exported as a CSV file named \texttt{hog\_features.csv}. This stored feature set served as a structured and compressed representation of the input images, which was later used for integration with deep learning models in the classification pipeline.

Overall, this HOG-based feature extraction step aimed to leverage handcrafted descriptors that capture geometrical and textural cues in retinal images, providing a complementary signal to the abstract features learned by convolutional layers in deep neural networks.

\subsection{Hybrid HOG-CNN Model}

The following section outlines the methodology for combining handcrafted gradient-based features with deep convolutional features to construct a hybrid classification model for retinal image analysis. This dual-branch framework is designed to support both binary and multi-class image classification tasks, effectively integrating traditional HOG descriptors with modern CNN-based feature representations.

In the preprocessing stage, each fundus image is formatted as a 3D tensor $\mathbf{I}_{\text{raw}} \in \mathbb{R}^{H \times W \times C}$, where $H$ and $W$ represent the image height and width, and $C = 3$ corresponds to the RGB color channels. No normalization is applied to pixel intensities; hence, the image tensor is passed directly into the model as:
\[
\mathbf{I}_{\text{input}} = \mathbf{I}_{\text{raw}}.
\]
All images are resized to a fixed dimension of $(224 \times 224)$ and grouped into batches of size $B$, represented as:
\[
\mathbf{X} = \{\mathbf{I}_1, \mathbf{I}_2, \ldots, \mathbf{I}_B\}, \quad \mathbf{y} = [y_1, y_2, \ldots, y_B],
\]
where $\mathbf{y}$ contains the ground-truth labels for classification.

In parallel with the RGB image input stream, each fundus image is converted to grayscale and processed using the Histogram of Oriented Gradients (HOG) algorithm to extract structural edge and texture information. This results in a HOG feature vector $\mathbf{H}_i \in \mathbb{R}^d$ for each image, where $d$ is the dimensionality of the HOG descriptor. These descriptors are passed through a three-layer fully connected neural network for dimensionality reduction and transformation. The first dense layer contains 800 ReLU-activated units, the second dense layer contains 256 ReLU-activated units, and the second contains 128 ReLU-activated units, yielding a HOG embedding vector:
\[
\mathbf{F}_{\text{HOG}} = \text{ReLU}(\mathbf{W}_2 \cdot \text{ReLU}(\mathbf{W}_1 \cdot \mathbf{H} + \mathbf{b}_1) + \mathbf{b}_2).
\]

Simultaneously, the RGB image $\mathbf{I}_i$ is input into a pre-trained convolutional neural network (CNN) such as ResNet-50 or EfficientNet. The classification head of the pre-trained model is removed by setting \texttt{include\_top = False}, and the convolutional base is frozen to preserve previously learned visual representations. The extracted feature maps are further processed by a 2D convolutional layer with 64 filters of size $(3 \times 3)$ and ReLU activation, followed by a max pooling layer with a $(2 \times 2)$ window. The output is then flattened and passed through a dense layer with 64 ReLU-activated units to produce the image-based embedding:
\[
\mathbf{F}_{\text{CNN}} \in \mathbb{R}^{64}.
\]

The two feature vectors $\mathbf{F}_{\text{CNN}}$ and $\mathbf{F}_{\text{HOG}}$ are concatenated to form a unified representation:
\[
\mathbf{F}_{\text{concat}} = [\mathbf{F}_{\text{CNN}}, \mathbf{F}_{\text{HOG}}] \in \mathbb{R}^{192}.
\]
This combined feature vector is then passed through two fully connected layers with 256 and 128 ReLU-activated units, respectively. A dropout layer with a dropout rate of $p = 0.2$ is applied to prevent overfitting. The final classification layer adapts to the task type. For binary classification, the output is computed using a sigmoid activation:
\[
\hat{y} = \text{Sigmoid}(\mathbf{W}_{\text{out}} \cdot \mathbf{F}_{\text{concat}} + \mathbf{b}_{\text{out}}),
\]
while for multi-class classification, a softmax activation is used:
\[
\hat{\mathbf{y}} = \text{Softmax}(\mathbf{W}_{\text{out}} \cdot \mathbf{F}_{\text{concat}} + \mathbf{b}_{\text{out}}).
\]

The model is trained to minimize the categorical cross-entropy loss:
\[
\mathcal{L} = -\frac{1}{B} \sum_{i=1}^{B} \sum_{c=1}^{C} y_{i,c} \log(\hat{y}_{i,c}),
\]
where $B$ is the batch size and $C$ is the number of output classes. Optimization is performed using the Adam optimizer. The model parameters $\theta$ are updated at each iteration according to:
\[
\theta_{t+1} = \theta_t - \eta \cdot \nabla_\theta \mathcal{L},
\]
where $\eta$ is the learning rate.

During inference, the predicted label for each image is determined by selecting the class with the highest predicted probability:
\[
\hat{y}_i = \arg\max_c \hat{y}_{i,c}.
\]
To evaluate the model, standard metrics are employed. These include \textbf{accuracy}, which reflects the proportion of correct predictions, \textbf{precision}, which measures the reliability of positive predictions, \textbf{recall} or sensitivity, which assesses how well actual positives are identified, and the \textbf{F1-score}, which balances precision and recall. Additionally, the \textbf{AUC} (Area Under the Receiver Operating Characteristic Curve) is used to evaluate the model's discriminative capability. All experiments are conducted using TensorFlow/Keras with CPU acceleration, with training performed over 50 epochs and a batch size of 32. The proposed HOG-CNN model demonstrates how combining engineered features with learned deep features can yield robust classification performance, especially in medical image analysis where both texture and contextual cues are important. The HOG-CNN architecture is visually illustrated in Figure~\ref{fig:HOG-CNN}, and the complete algorithmic steps are summarized in here ~\ref{alg:hog_cnn_model}.

\begin{algorithm}
\SetAlgoNlRelativeSize{0}
\DontPrintSemicolon
\caption{Training Procedure for Hybrid HOG-CNN Model}
\label{alg:hog_cnn_model}

\KwIn{Fundus image dataset $\mathcal{D} = \{(\mathbf{I}_i, y_i)\}_{i=1}^{N}$, classification type (binary or multi-class), number of epochs $E$}
\KwOut{Trained HOG-CNN model and evaluation metrics}

\textbf{Step 1: HOG Feature Extraction} \\
\For{$i \gets 1$ \KwTo $N$}{
    Convert $\mathbf{I}_i$ to grayscale\;
    Extract HOG descriptor $\mathbf{H}_i \in \mathbb{R}^d$\;
    Pass $\mathbf{H}_i$ through Dense(800, ReLU) $\rightarrow$ Dense(256, ReLU) $\rightarrow$ Dense(128, ReLU) to obtain HOG embedding $\mathbf{F}_{\text{HOG}}$\;
}

\textbf{Step 2: CNN Feature Extraction} \\
\For{$i \gets 1$ \KwTo $N$}{
    Resize $\mathbf{I}_i$ to $(224,224,3)$ without normalization\;
    Pass $\mathbf{I}_i$ through pre-trained CNN backbone (with frozen weights, \texttt{include\_top=False})\;
    Apply Conv2D(64 filters, $3\times3$, ReLU), MaxPooling($2\times2$), Flatten, Dense(64, ReLU) to obtain CNN embedding $\mathbf{F}_{\text{CNN}}$\;
}

\textbf{Step 3: Feature Fusion and Classification} \\
\For{$i \gets 1$ \KwTo $N$}{
    Concatenate $\mathbf{F}_{\text{CNN}}$ and $\mathbf{F}_{\text{HOG}}$ to get $\mathbf{F}_{\text{concat}} \in \mathbb{R}^{192}$\;
    Pass $\mathbf{F}_{\text{concat}}$ through Dense(256, ReLU) $\rightarrow$ Dense(128, ReLU) $\rightarrow$ Dropout(0.2)\;
    Apply output layer with Sigmoid (binary) or Softmax (multi-class) activation to get predictions $\hat{\mathbf{y}}$\;
}

Train the model over $50$ epochs minimizing categorical cross-entropy loss using Adam optimizer; evaluate with Accuracy, Precision, Recall, F1-score, AUC metrics.

\Return{Trained HOG-CNN model and evaluation metrics}
\end{algorithm}

\begin{figure*}[t!] 
	\centering
	\subfloat[\scriptsize Training vs. Test Accuracy (Binary Classification)\label{fig:Train-Test Acc img}]{%
		\includegraphics[width=0.45\linewidth]{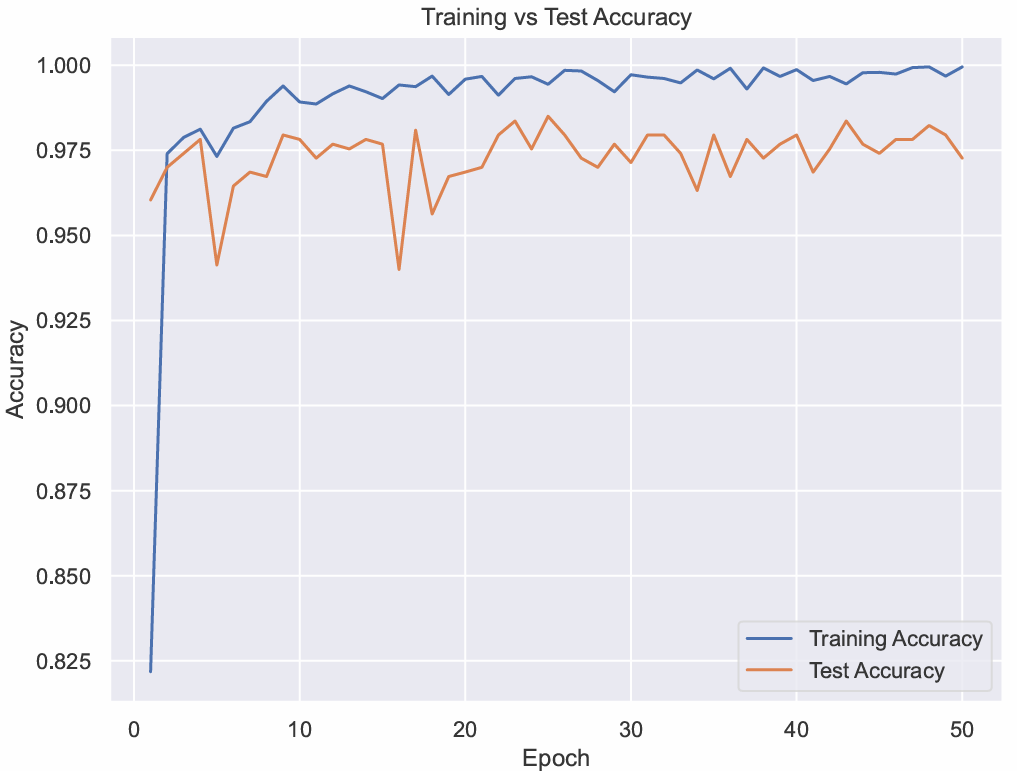}}
	\hspace{0.05\linewidth}
	\subfloat[\scriptsize Training vs. Test Loss (Binary Classification)\label{fig:Train-Test Loss img}]{%
		\includegraphics[width=0.45\linewidth]{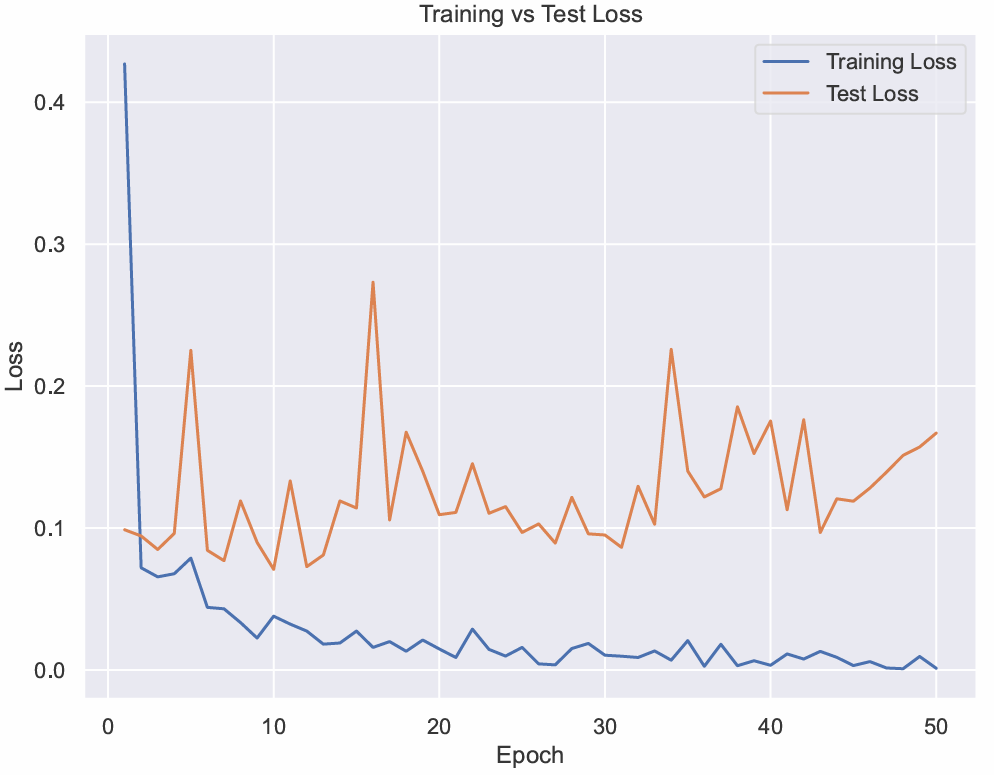}}
	\caption{\footnotesize Training and Testing Accuracy and Loss over 50 Epochs on the APTOS Dataset (Binary Classification).}
	\label{fig:Acc-Loss}
\end{figure*}

\section{Experiments}
\label{sec:experiments}

\subsection{Datasets} \label{sec:datasets}

To see the performance of our Hog-ML model for Glaucoma, DR,  and AMD screening, we did several experiments on well-known benchmark datasets. We give the basic details of these datasets in \Cref{tab:datasets}. 
Further details (resolution, camera, etc.) for all the datasets can be found in~\cite{li2021applications}. 

\begin{table}[b]
\centering
\caption{ Benchmark datasets for fundus images. \label{tab:datasets}}
\setlength\tabcolsep{2 pt}
\small
\begin{tabular}{lcccc} 
\toprule
\textbf{Dataset} & \textbf{Disease} &  \textbf{Total} & \textbf{Normal} & \textbf{Abnormal}   \\
\midrule
APTOS 2019
& DR & 3662 & 1805 & 1857   \\
ORIGA  
& Gl &  650 & 482 & 168   \\

IChallenge-AMD 
& AMD & 400 & 311 & 89  \\

\bottomrule
\end{tabular} 
\end{table}

\smallskip

\noindent {\bf IChallenge-AMD dataset} is 
designed for the Automatic Detection challenge on Age-related Macular  degeneration (ADAM Challenge) which was held as a satellite event of the ISBI 2020 conference~\cite{fang2022adam, ichallenge-amd}. 
There are two different resolutions of images, i.e., $2124\times2056$ pixels (824 images) and $1444\times 1444$ (376 images). While the dataset has 1200 images, only 400 of them are available with labels. Like most other references, we used these 400 images in our experiments~\Cref{tab:amd}. Among these 400 images, 89 of them are labeled as AMD, and the remaining 311 images are labeled as healthy.

\smallskip

\noindent {\bf ORIGA dataset} contains 650 high resolution ($3072\times 2048$) retinal images for Glaucoma annotated by trained professionals from Singapore Eye Research Institute~\cite{zhang2010origa}. Out of 650 fundus images, 168 images are labeled as Glaucoma and the remaining 482 images are labeled as healthy. 

\smallskip

\noindent {\bf APTOS 2019 dataset} was used for a Kaggle competition on DR diagnosis~\cite{aptos2019}. The images have varying resolutions, ranging from $474\times 358$ to $3388\times 2588$. APTOS stands for Asia Pacific Tele-Ophthalmology Society, and the dataset was provided by Aravind Eye Hospital in India. 
In this dataset, fundus images are graded manually on a scale of 0 to 4 (0: no DR; 1: mild; 2: moderate; 3: severe; and 4: proliferative DR) to indicate different severity levels. The number of images in these classes are respectively 1805, 370, 999, 193, and 295. In the binary setting, class 0 is defined as the normal group, and the remaining classes (1-4) are defines as DR group which gives a split 1805:1857. The total number of training and test samples in the dataset were 3662 and 1928 respectively. However, the labels for the test samples were not released after the competition, so like other references, we used the available 3662 fundus images with labels. We report our results on a binary and 5-class classification setting. In a binary setting,  fundus  images with grades 1, 2, 3, and 4 are identified as DR group, and grade 0 images as the normal group.

\subsection{Experimental Setup} \

\noindent {\bf Training–Test Split:}  
Since none of the datasets provide a predefined \textit{train:test} split, prior works have adopted varying strategies for data partitioning.  
To ensure fair comparison, we follow the most commonly used splits reported in the literature.  
Specifically, for the APTOS dataset (both binary and 5-class tasks) and the ORIGA dataset, we adopt an 80:20 training-to-testing ratio for evaluating our HOG-CNN model.  
For the IChallenge-AMD dataset, we employ 5-fold cross-validation and report the mean performance across all folds.
 
Because of the discrepancy between the experimental setups of different methods, we give the train:test splits of all models in our accuracy tables to facilitate a fair comparison (\Cref{tab:amd,tab:aptos,tab:origa,tab:aptos_multi}). 
\smallskip

\noindent {\bf No Data Augmentation:} Note that as our datasets are quite small and imbalanced compared to other image classification tasks for deep learning models, hence all CNN and other deep learning methods need to use serious data augmentation (sometimes 50-100 times) to train their model and avoid overfitting~\cite{goutam2022comprehensive}. We do not use any type of data augmentation or pre-processing to increase the size of training data for HOG-CNN model. This makes our model computationally very efficient. We used the original datasets and did not use any kind of data augmentation as we used pre-trained models as backbone.

\smallskip


\smallskip

\noindent {\bf HOG-CNN Model Hyperparameters:} We extracted $26{,}244$ HOG features per image. Then we integrated these features with pre-trained CNN features. We trained the HOG-CNN model for 50 epochs using a batch size of 32 for all datasets. We used the Adam optimizer and kept the remaining parameters as default. To reduce overfitting, we incorporated a Dropout layer with a rate of 0.2 and employed early stopping with a patience of 10, restoring the best-performing model weights for final prediction.

\smallskip

Our code is available at the following link~\footnote{ \url{https://github.com/FaisalAhmed77/HOG-CNN}}. 


\vspace{-.1in}

\begin{figure*}[t!]
    \centering
    \subfloat[\scriptsize APTOS Binary Classification Accuracy Comparison\label{fig:APTOS_Acc_Binary}]{%
        \includegraphics[width=0.31\linewidth]{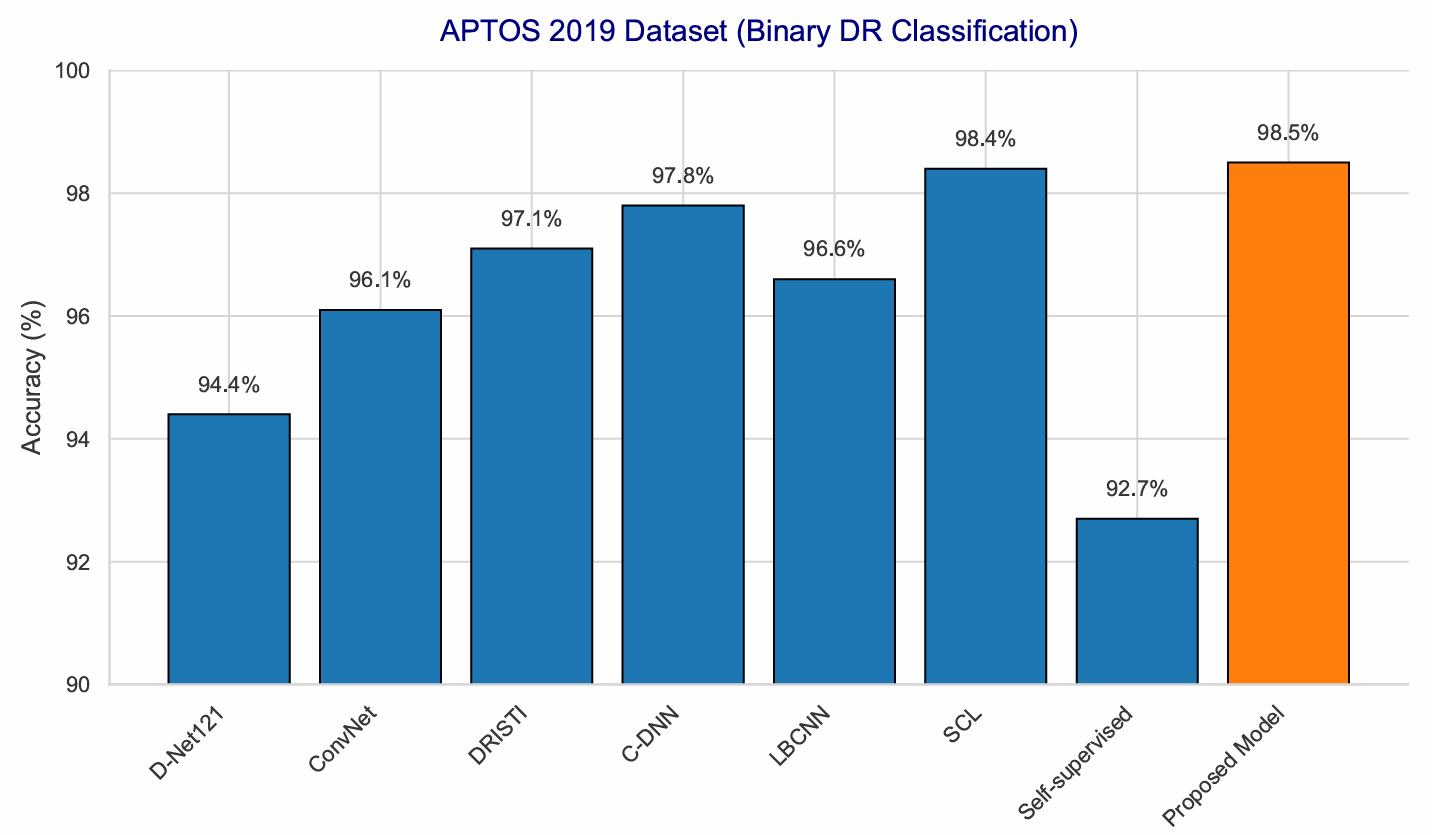}}
    \hspace{0.03\linewidth}
    \subfloat[\scriptsize APTOS 5-Class Classification Accuracy Comparison\label{fig:APTOS_Acc_5Class}]{%
        \includegraphics[width=0.31\linewidth]{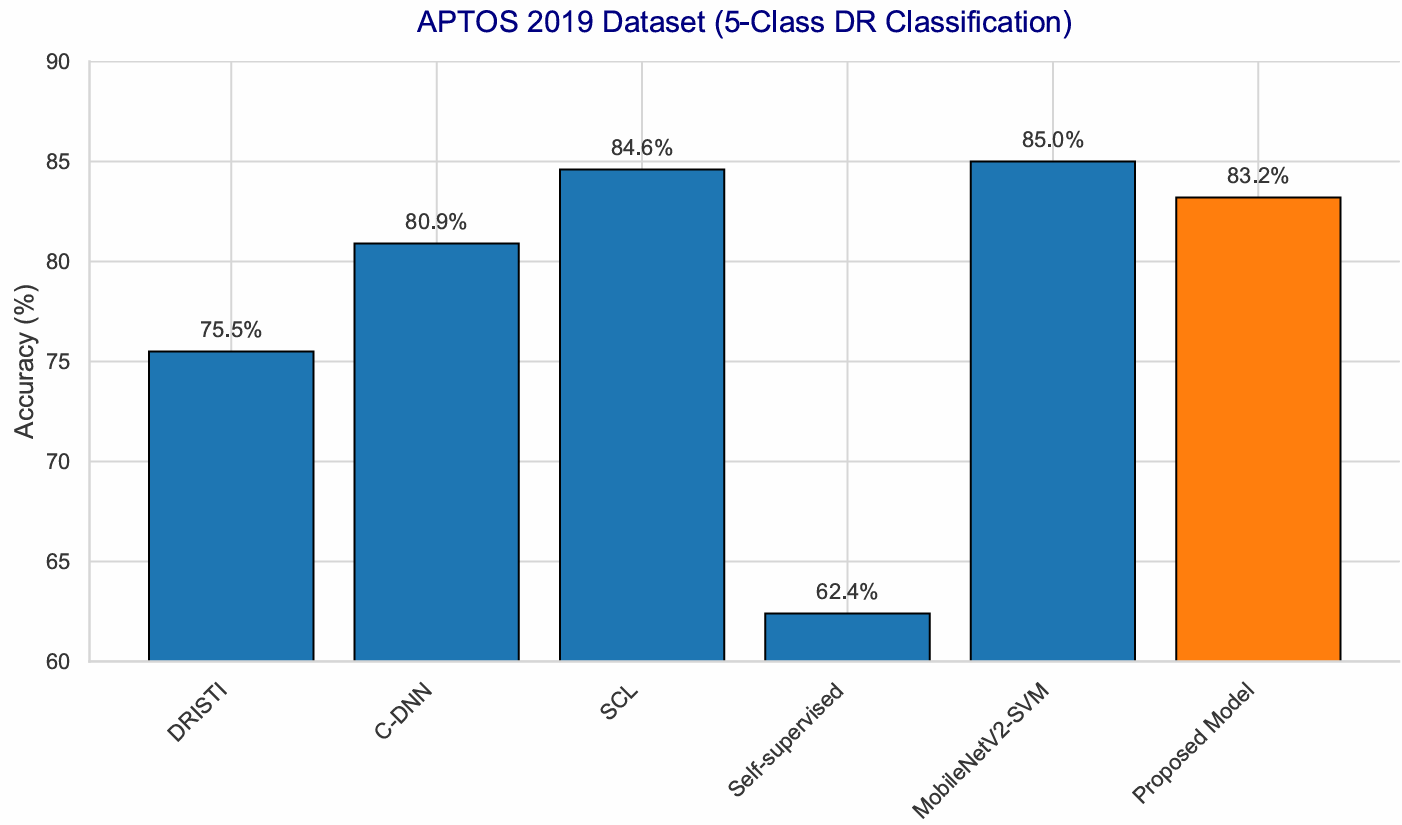}}
    \hspace{0.03\linewidth}
    \subfloat[\scriptsize IC-Dataset Model Performance Comparison (Radar Plot)\label{fig:IC_Acc_Spider}]{%
        \includegraphics[width=0.31\linewidth]{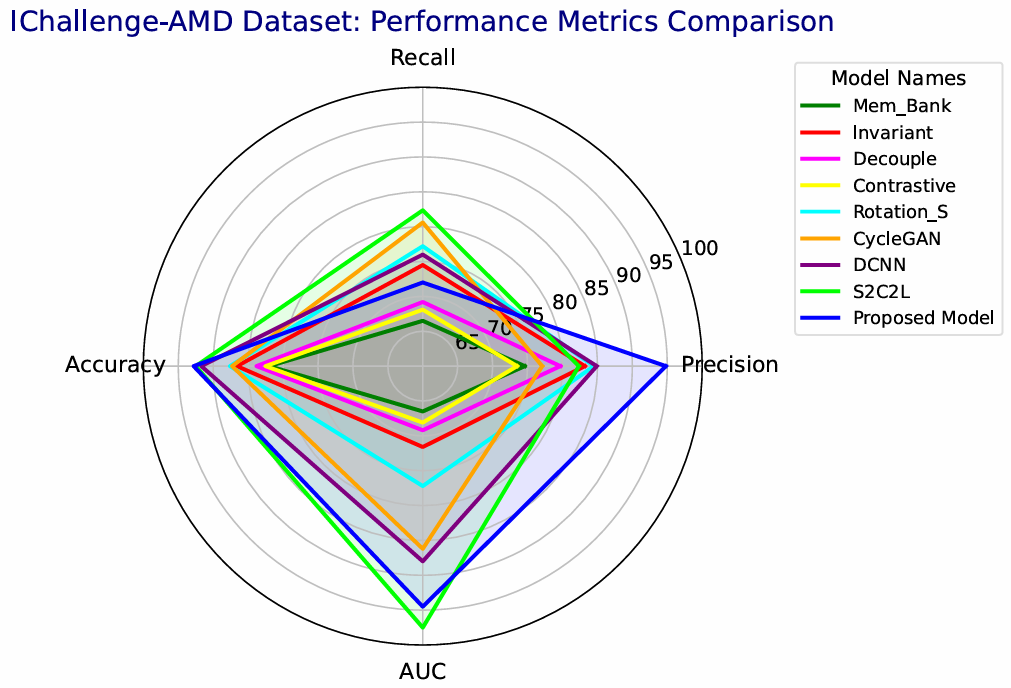}}
    \caption{\footnotesize Accuracy comparison of the proposed model with existing deep learning methods: (a) and (b) show bar plots for binary and 5-class classification on the APTOS 2019 dataset, respectively; (c) presents a radar chart comparing multiple metrics on the IC-Dataset.}
    \label{fig:Acc_Comparison}
\end{figure*}

\section{Results} \label{sec:accuracy} 

\noindent {\bf AMD Detection Results:} 
Table~\ref{tab:amd} presents the classification performance of recent methods on the \textit{IChallenge-AMD} dataset.  
Our proposed HOG-CNN attains the highest overall accuracy of 92.8\%, surpassing the previous best DCNN (91.7\%) and edging out the self-supervised S2C2L (92.5\%).  
HOG-CNN also records the best precision (94.8\%), indicating a notably low false-positive rate.  
Conversely, S2C2L achieves the top recall (82.4\%) and the highest AUC (97.5\%), showing stronger sensitivity but slightly lower specificity than our model.  
Other self-supervised approaches—including Rotation S (87.6\% Acc), Invariant (86.6\% Acc), and Mem-Bank, Decouple, and Contrastive (all $\le$ 83.8\% Acc)—trail behind both HOG-CNN and S2C2L.  
These results demonstrate that integrating handcrafted HOG descriptors with deep CNN representations yields more discriminative features, delivering superior AMD detection accuracy compared with existing purely CNN-based or self-supervised techniques.

\noindent {\bf DR Detection Results:} We report diabetic retinopathy (DR) detection results using the APTOS 2019 dataset under both binary and multiclass classification settings, as presented in \Cref{tab:aptos} and \Cref{tab:aptos_multi}, respectively. In the binary classification setting (\Cref{tab:aptos}), our proposed HOG-CNN model achieved the best overall performance, obtaining a precision, recall, and accuracy of 98.5\%, along with an AUC of 99.2\%. This outperforms several state-of-the-art models, including SCL~\cite{islam2022applying} (98.4\% accuracy, 98.9\% AUC), C-DNN~\cite{bodapati2021composite} (97.8\% accuracy), and LBCNN~\cite{macsik2022local} (96.6\% accuracy, 98.7\% AUC). Notably, our method surpasses these models in all reported metrics, indicating superior diagnostic capability and a stronger balance between sensitivity and specificity.

In the more challenging multiclass classification setting (\Cref{tab:aptos_multi}), which involves classifying DR into five severity levels, the HOG-CNN model continued to demonstrate competitive performance. It achieved a precision of 86.0\%, recall of 78.6\%, accuracy of 83.2\%, and the highest AUC of 94.2\% among all compared methods. While MobileNetV2-SVM~\cite{singh2024mobilenetv2} slightly outperformed in terms of accuracy (85.0\%) and recall (74.8\%), our model offered the best precision and AUC, making it more reliable for nuanced multiclass predictions. Other approaches, such as SCL~\cite{islam2022applying} (84.6\% accuracy, 93.8\% AUC) and C-DNN~\cite{bodapati2021composite} (80.9\% accuracy), lagged behind in overall metrics.

These results demonstrate that the proposed integration of handcrafted HOG features with deep CNN representations enhances the model’s ability to capture both local texture variations and high-level semantic features, leading to superior performance in both binary and multiclass DR diagnosis tasks. The consistently high AUC scores also suggest that our HOG-CNN model maintains strong discriminative power across different DR severity levels, making it a promising tool for real-world clinical applications.

\noindent {\bf Glaucoma Detection Results:} We evaluate the performance of our HOG-CNN model for glaucoma diagnosis using the ORIGA dataset, as summarized in \Cref{tab:origa}. The proposed model achieved an accuracy of 83.9\%, sensitivity of 83.9\%, specificity of 82.0\%, and an AUC of 87.2\%. Although the performance does not surpass the top-performing models such as CAD~\cite{singh2024three} (96.5\% accuracy, 98.1\% sensitivity, 94.2\% AUC) and ODGNet~\cite{latif2022odgnet} (95.8\% accuracy, 94.8\% sensitivity, 97.9\% AUC), our model still shows solid results given its simplicity and lack of domain-specific pretraining or segmentation-based preprocessing.

In contrast to methods like EAMNet~\cite{liao2019clinical} and 18-CNN~\cite{elangovan2021glaucoma}, which either did not report full performance metrics or underperformed in terms of sensitivity (e.g., 58.1\% for 18-CNN), our HOG-CNN model strikes a better balance between sensitivity and specificity. It also compares competitively to SVM-SMOTE~\cite{zhao2019glaucoma}, which achieved 82.8\% accuracy and 88.9 AUC, slightly lower than our AUC of 87.2.

While ensemble or segmentation-based models like CAD~\cite{singh2024three}, ODGNet~\cite{latif2022odgnet}, and CNN-SVM~\cite{ajitha2021identification} achieve superior performance, they typically rely on complex pipelines, including optic disc segmentation or pretrained architectures. In contrast, our model integrates handcrafted HOG features with CNN-based representations in a lightweight and efficient framework. These findings support the utility of the HOG-CNN model as a viable and generalizable approach, especially in scenarios with limited computational resources or where interpretability of feature representations is important.

\begin{table}[t]
\centering
\caption{ Accuracy results for AMD diagnosis. \label{tab:amd}}
\setlength\tabcolsep{3pt}
\footnotesize
\resizebox{1.\linewidth}{!}{
\begin{tabular}{lccccccc}
\multicolumn{8}{c}{\bf{IChallenge-AMD Dataset}} \\
\toprule
\textbf{Method} & \textbf{Nor:Abn} & \textbf{Train:Test} & \textbf{Class}  &\textbf{Prec} &\textbf{Recall}& \textbf{Acc}  & \textbf{AUC} \\
\midrule
Mem-Bank~\cite{wu2018unsupervised} &  311:89 & 5 fold CV & 2&74.6&66.5&82.0 & 66.5 \\
Invariant~\cite{ye2019unsupervised} &  311:89 & 5 fold CV & 2&83.2&74.5&  86.6 & 71.6   \\
Decouple~\cite{feng2019self} &  311:89 & 5 fold CV & 2&79.7&69.2&  83.8 & 69.2 \\
Contrastive~\cite{chen2020simple} &  311:89 & 5 fold CV & 2&73.5&68.1& 82.5 & 68.1  \\
Rotation S~\cite{li2021rotation} &  311:89 & 5 fold CV & 2&84.5&77.2&  87.6 & 77.2  \\
CycleGAN~\cite{zhang2021joint}& 933:267 & 5 fold CV & 2 & 77.1& \underline{80.6}&87.3&86.2 \\
DCNN~\cite{chakraborty2022dcnn} & 311:89 & 10 fold CV & 2 & \underline{84.9} & 76.0 & 91.7 & 88.0   \\

S2C2L ~\cite{bi2024self} & 311:89 & 5 fold CV & 2 & 82.35 &\textbf{82.35} &\underline{92.50} & \textbf{97.5}\\

\midrule

\bf{HOG-CNN} &   311:89 &  5 fold CV &2& \textbf{94.8} & 72.0 & \textbf{92.8} & \underline{94.5}  \\
\bottomrule
\end{tabular}}
\end{table}

\begin{table}[t]
\centering
\caption{Accuracy results for binary DR diagnosis. \label{tab:aptos}}
\vspace{.1cm}
\setlength\tabcolsep{3 pt}
\footnotesize
\resizebox{1.\linewidth}{!}{
\begin{tabular}{lccccccc}
\multicolumn{8}{c}{\bf{APTOS 2019 Dataset Binary (DR)}} \\
\toprule
\textbf{Method} & \textbf{Nor:Abn} & \textbf{Train:Test} & \textbf{Class}  &\textbf{Prec} &\textbf{Recall} & \textbf{Acc} & \textbf{AUC}  \\
\midrule
D-Net121~\cite{chaturvedi2020automated} &  1805:1857 & 85:15 & 2&86.0&87.0&94.4  & -\\
ConvNet~\cite{bodapati2020blended} &  1805:1857 & 80:20 & 2&-& - &96.1   & - \\
DRISTI~\cite{kumar2021dristi} &  1805:1857 & 85:15 & 2&-& - &97.1    & - \\
C-DNN~\cite{bodapati2021composite} &  1805:1857 & 85:15 & 2&98.0& 98.0 &97.8   & - \\
LBCNN~\cite{macsik2022local} & 1805:1857 & 80:20 &  2 & - & - & 96.6 &98.7\\
SCL~\cite{islam2022applying} &  1805:1857 & 85:15 & 2&\underline{98.4}&\underline{98.4}&\underline{98.4} & \underline{98.9}\\
Self-supervised \cite{long2024classification}&  1805:1857 & 10 fold & 2 &- &- &92.7 & -\\
\midrule
\bf{HOG-CNN} &   1805:1857 & 80:20 &2&\textbf{98.5}  &\textbf{98.5}&\textbf{98.5}&  \textbf{99.2}\\
\bottomrule
\end{tabular}}
\end{table}

\begin{table}[h!]
\centering
\caption{Accuracy results for multiclass DR diagnosis. \label{tab:aptos_multi}}
\setlength\tabcolsep{4 pt}
\footnotesize
\resizebox{\linewidth}{!}{
\begin{tabular}{lccccccc}
\multicolumn{8}{c}{\bf{APTOS 2019 Dataset - 5-class (DR)}} \\
\toprule
\textbf{Method} & { \textbf{Nor:Abn}} & \textbf{Train:Test} &  \textbf{Class}  &\textbf{Prec} &\textbf{Rec} & \textbf{Acc} & \textbf{AUC}  \\
\midrule
DRISTI~\cite{kumar2021dristi} &  1805:1857 & 85:15 & 5 &59.4  &54.6 &75.5  & -\\
C-DNN~\cite{bodapati2021composite} &  1805:1857 & 85:15 &  5 & -&- &80.9 &-\\
SCL~\cite{islam2022applying} &  1805:1857 & 85:15 & 5 &\underline{73.8} &70.50 &\underline{84.6} & \underline{93.8}\\
Self-supervised \cite{long2024classification}&  1805:1857 & 10 fold & 5 &- &- &62.4 & -\\
MobileNetV2-SVM \cite{singh2024mobilenetv2}&  1805:1857 & - & 5 &72.0 &\underline{74.8} &\textbf{85.0} & 92.0\\
\midrule
\bf{HOG-CNN} &   1805:1857 & 80:20 &5& \textbf{86.0}& \textbf{78.6}& 83.2 & \textbf{94.2}\\
\bottomrule
\end{tabular}}
\vspace{-.1in}
\end{table}

\begin{table}[t]
\centering
\caption{Accuracy results for Glaucoma diagnosis. \label{tab:origa}}
\vspace{.1cm}
\setlength\tabcolsep{3 pt}
\footnotesize
\resizebox{1.\linewidth}{!}{
\begin{tabular}{lccccccc}
\multicolumn{8}{c}{\bf{ORIGA Dataset (Glaucoma)}} \\
\toprule
\textbf{Method} & \textbf{Nor:Abn} & \textbf{Train:Test} & \textbf{Class} &\textbf{Sen}&\textbf{Spec}& \textbf{Acc}  & \textbf{AUC} \\
\midrule
EAMNet~\cite{liao2019clinical} &  482:168 & 2 fold CV & 2&-&-&  - & 88.0  \\
SVM-SMOTE \cite{zhao2019glaucoma} & 482:168 & 10 fold CV & 2&87.6 & 77.9 & 82.8 & 88.9 \\
18-CNN~\cite{elangovan2021glaucoma} & 482:168 & 70:30 & 2 & 58.1 & 92.4 & 78.3 & - \\
NasNet~\cite{taj2021ensemble} & 482:168 & 70:30 & 2 & 78.7 & 91.1 & 87.9 & - \\
CNN-SVM~\cite{ajitha2021identification} & 660:453 & 70:30 & 2 & 89.5 & \textbf{100} &95.6 & - \\
ODGNet~\cite{latif2022odgnet} &  482:168 & pretrained &2&\underline{94.8}&\underline{94.9}& \underline{95.8} & \textbf{97.9}  \\
CAD\cite{singh2024three} &  482:168 & 10 fold CV &2&\textbf{98.10}&93.3& \textbf{96.50} & \underline{94.2}  \\
\midrule

\bf{HOG-CNN} &   482:168 & 80:20 &2& 83.9 & 82.0 &83.9 & 87.2  \\
\bottomrule
\end{tabular}}
\end{table}


\section{Discussion} \label{sec:discussion}

This study comprehensively evaluated the proposed HOG-CNN model across three major retinal disease diagnosis tasks Age-related Macular Degeneration (AMD), Diabetic Retinopathy (DR), and Glaucoma demonstrating consistently competitive performance against state-of-the-art baselines. For AMD detection on the IC-AMD dataset, the proposed model achieved an accuracy, precision, and recall of 92.8\%, along with an AUC of 94.5\%, outperforming most existing methods. While S2C2L~\cite{bi2024self} attained a slightly higher AUC of 97.5\%, our model offers a more balanced performance across all metrics, demonstrating robust and consistent classification capability.
 In DR detection using the APTOS 2019 dataset, HOG-CNN delivered outstanding results across both binary and multiclass settings. It outperformed competing models in binary classification with 98.5\% precision, recall, and accuracy, and a leading AUC of 99.2\%. In the multiclass task, it maintained high performance (83.2\% accuracy, 94.2\% AUC), demonstrating reliability for more granular DR grading. These results emphasize the model’s strong discriminative power and adaptability across varied classification challenges.

In glaucoma detection using the ORIGA dataset, HOG-CNN achieved an accuracy of 83.9\%, sensitivity of 83.9\%, specificity of 82.0\%, and an AUC of 87.2\%. While it did not surpass ensemble or segmentation-heavy models like CAD~\cite{singh2024three} or ODGNet~\cite{latif2022odgnet}, its performance remains competitive, especially given its simplicity and efficiency. Notably, the model requires no domain-specific pretraining or segmentation, offering a practical advantage in low-resource settings. Overall, the integration of handcrafted HOG features with CNN-based deep representations enables a lightweight yet powerful framework, capable of capturing both fine-grained textures and high-level semantics. These findings validate the HOG-CNN as a generalizable and interpretable tool for retinal disease screening in real-world clinical applications.


\section{Limitations} \label{sec:limitations}

Although the proposed HOG‑CNN framework delivers state‑of‑the‑art or near‑state‑of‑the‑art results across AMD, DR, and glaucoma benchmarks, its reliance on handcrafted HOG descriptors makes performance sensitive to image resolution and illumination, limits the value of standard geometric data augmentation, and may exclude richer, fully learnable representations; moreover, the off‑the‑shelf CNN backbone is not fine‑tuned to the ophthalmic domain, and evaluation on the small ORIGA dataset constrains the generalisability of glaucoma findings, highlighting the need for larger, heterogeneous datasets and task‑specific end‑to‑end training in future work.

\section{Conclusion} \label{sec:conclusion}

This study introduces a hybrid HOG-CNN framework for retinal disease diagnosis, achieving strong and consistent results across AMD, DR, and glaucoma classification tasks. By fusing handcrafted Histogram of Oriented Gradients (HOG) features with deep convolutional representations, the model captures both fine-grained texture and high-level semantic information. On the IC-AMD dataset, the proposed model achieved an accuracy of 92.8\% and an AUC of 94.5\%, surpassing several state-of-the-art methods in overall performance. In binary DR detection using the APTOS 2019 dataset, it attained 98.5\% accuracy and 99.2\% AUC, surpassing leading models such as SCL and C-DNN. For multiclass DR classification, it delivered 83.2\% accuracy and the highest AUC of 94.2\%, indicating robust performance in nuanced clinical grading. While performance on the ORIGA dataset for glaucoma (83.9\% accuracy, 87.2\% AUC) was slightly below the top-performing models, the HOG-CNN still demonstrated competitive results given its simplicity and lack of specialized preprocessing.

The model’s lightweight and interpretable architecture makes it well-suited for deployment in real-world, resource-limited settings. Its ability to outperform or match more complex models—without requiring extensive augmentation, segmentation, or transfer learning—underscores the value of integrating handcrafted and learned features. These findings highlight HOG-CNN as a practical and generalizable approach for automated retinal disease screening and a promising foundation for future clinical decision support systems.


\section{Future Work} \label{sec:future_work}

Future work will focus on fine-tuning the CNN backbone with domain-specific data, integrating more expressive topological features beyond HOG, and developing robust normalization methods to handle resolution and illumination variability. Expanding evaluation to larger, diverse datasets and real-world clinical settings is also essential. Additionally, incorporating interpretability tools tailored to hybrid models can enhance transparency and support clinical adoption.



\section*{Declarations}

\textbf{Funding} \\
The author received no financial support for the research, authorship, or publication of this work.

\textbf{Acknowledgement} \\
The authors utilized an online platform to check and correct grammatical errors and to improve sentence readability.

\vspace{2mm}
\textbf{Conflict of interest/Competing interests} \\
The authors declare no conflict of interest.

\vspace{2mm}
\textbf{Ethics approval and consent to participate} \\
Not applicable. This study did not involve human participants or animals, and publicly available datasets were used.

\vspace{2mm}
\textbf{Consent for publication} \\
Not applicable.

\vspace{2mm}
\textbf{Data availability} \\
The datasets used in this study are publicly available. The APTOS 2019 dataset is accessible at \url{https://www.kaggle.com/c/aptos2019-blindness-detection}, and the IChallenge-AMD dataset was provided for the ADAM challenge as part of ISBI 2020.

\vspace{2mm}
\textbf{Materials availability} \\
Not applicable.

\vspace{2mm}
\textbf{Code availability} \\
The source code used in this study is publicly available at \url{https://github.com/FaisalAhmed77/HOG-CNN}.

\vspace{2mm}
\textbf{Author's Contribution} \\
FA conceptualized the study, downloaded the data, prepared the code, performed the data analysis and wrote the manuscript. FA reviewed and approved the final version of the manuscript. 

\clearpage

\section{APPENDIX}
\begin{appendix}

Below, we provide additional performance metrics for our HOG-CNN model, utilizing various pre-trained CNN models as backbones.

\begin{table}[h!]
\centering
\caption{Accuracy results for HOG-CNN on APTOS dataset. \label{tab:aptos_HOG_CNN}}
\setlength\tabcolsep{10 pt}
\begin{tabu}{lcccc}
\multicolumn{5}{c}{\bf{APTOS 2019 Dataset (DR) - Binary}} \\
\toprule
\textbf{Method} & \textbf{Acc} &\textbf{Prec} & \textbf{Rec} & \textbf{AUC}  \\
\midrule
Resnet50 &96.73 &96.73 &96.73&  98.28\\
\rowfont{\color{blue}}
HOG + Resnet50&98.09&97.96&98.09&98.54\\
\hline                               
DenseNet201 &96.55&96.72&96.55&97.81 \\
\rowfont{\color{blue}}

HOG + DenseNet201 &96.45 &96.45 &96.45 &98.14 \\
    \hline
InceptionResNetV2 &52.91 &52.91&52.91 &52.91 \\
\rowfont{\color{blue}}

HOG + InceptionResNetV2 &94.54 &94.54 &94.41 &94.72 \\
\hline
MobileNetV2 &94.73 &94.73 &94.73 &96.72\\
\rowfont{\color{blue}}

HOG + MobileNetV2 &95.50 &95.37 &95.50 &97.03\\
\hline
EfficientNetB3&96.91&97.09&96.91&98.68\\
\rowfont{\color{blue}}
HOG + EfficientNetB3&98.50&98.47&98.50&99.21 \\
\hline
Xception&94.73&94.74&94.91&97.47\\
\rowfont{\color{blue}}
HOG + Xception&94.68&94.68&94.68&93.03 \\
\hline
VGG19&96.91&96.91&96.91&97.80 \\
\rowfont{\color{blue}}
HOG + VGG19&96.18&96.31&96.04&97.97 \\
\hline
InceptionV3 &91.64 &91.64 &91.64&95.87 \\
\rowfont{\color{blue}}
HOG + InceptionV3 &95.36&95.36&95.36&95.85\\
\hline
EfficientNetB0 &97.64 &97.64 &97.64 &98.62\\
\rowfont{\color{blue}}
HOG + EfficientNetB0 &98.36&98.36&98.36&98.86\\
\hline
EfficientNetB2 &96.91 &96.90 &96.73&98.28 \\
\rowfont{\color{blue}}
HOG + EfficientNetB2&98.36 &98.36 &98.36&98.94 \\

\bottomrule
\end{tabu}
\end{table}

\begin{table*}[h!]
\centering
\caption{Accuracy results for our HOG-CNN models on APTOS dataset (5-labels) with different backbones. \label{tab:aptos_multi_HOG_CNN}}
\normalsize
\setlength\tabcolsep{8 pt}
\begin{tabu}{lcccc}
\multicolumn{5}{c}{\bf{APTOS 2019 Dataset (DR) for 5-class}} \\
\toprule
\textbf{Method} & \textbf{Acc} &\textbf{Prec} & \textbf{Rec} & \textbf{AUC}  \\
\midrule

Resnet50&76.73&77.67&75.27&90.54 \\
\rowfont{\color{blue}}
HOG + Resnet50&78.99&83.36&72.44&89.43  \\
    \hline                           
DenseNet201&74.36&75.83&70.18 &90.81 \\
\rowfont{\color{blue}}
HOG + DenseNet201&73.64&78.98 &70.36 &92.53 \\
\hline
MobileNetV2 &71.27&73.00 &69.82 &88.80 \\
\rowfont{\color{blue}}
HOG + MobileNetV2&78.17 &78.61 &77.22&80.89 \\
\hline
EfficientNetB3 &76.00&76.74 &74.36 &92.85 \\
\rowfont{\color{blue}}
HOG + EfficientNetB3 &82.26 &86.32 &77.49&93.66 \\
\hline
Xception &71.09&77.63 &64.36 &91.00 \\
\rowfont{\color{blue}}
HOG + Xception &79.13&82.15 &74.08 &88.80 \\
\hline
VGG19 &78.55&86.73&71.27&95.65 \\
\rowfont{\color{blue}}
HOG + VGG19 &79.67 &84.03 &71.76 &92.26 \\
\hline
InceptionV3 &69.45&69.91 &68.00&89.09 \\
\rowfont{\color{blue}}
HOG + InceptionV3&77.49 &78.04&76.13 &82.41 \\
\hline
EfficientNetB0&78.55&78.83&78.55&92.34 \\
\rowfont{\color{blue}}
HOG + EfficientNetB0&83.49 &84.09&82.95 &88.70 \\
\hline
EfficientNetB2 &78.55 &79.81 &76.91 &94.25 \\
\rowfont{\color{blue}}
HOG + EfficientNetB2&83.22&85.97&78.58 &94.15 \\

\bottomrule
\end{tabu}
\end{table*}

\begin{table*}[h!]
\centering
\caption{Accuracy results for our HOG-CNN models on IChallenge dataset with different backbones. \label{tab:ic_HOG_CNN}}
\normalsize
\setlength\tabcolsep{8 pt}
\begin{tabu}{lcccc}
\multicolumn{5}{c}{\bf{IChallenge Dataset (AMD)}} \\
\toprule
\textbf{Method} & \textbf{Acc} &\textbf{Prec} & \textbf{Rec} & \textbf{AUC}  \\
\midrule
Resnet50&87.34 &87.34&87.34 &86.54 \\
\rowfont{\color{blue}}                              
HOG + Resnet50&91.75&86.56 &76.27&92.01 \\
\hline                            
                               
DenseNet201&78.48&79.49&78.48&88.05  \\
 \rowfont{\color{blue}}
HOG + DenseNet201 &89.50&80.38 &70.92 &89.78 \\
\hline    
InceptionResNetV2 &21.52&2152 &21.52 &50.00 \\       
\rowfont{\color{blue}}
HOG + InceptionResNetV2&87.50 &87.34&86.25 &89.14 \\
\hline
MobileNetV2&86.08 &85.90 &84.81 &91.67 \\
\rowfont{\color{blue}}
HOG + MobileNetV2 &90.00 &88.89 &90.00 &92.23 \\ 
\hline 
EfficientNetB3 &88.61 &88.75 &89.87 &95.95 \\
\rowfont{\color{blue}}
HOG + EfficientNetB3 &92.75 &88.71&77.52 &93.61 \\
\hline 
Xception &84.81 &84.81 &84.81 &88.24 \\
\rowfont{\color{blue}}
HOG + Xception&86.25&87.84 &81.25 &91.11 \\
\hline 
VGG19 &86.08 &86.08 &86.08 &92.31 \\
\rowfont{\color{blue}}
HOG + VGG19 &90.00 &88.89 &90.00 &90.95 \\
\hline 
InceptionV3 &72.15 &72.15 &72.15 &82.73 \\
\rowfont{\color{blue}}
HOG + InceptionV3 &87.50 &87.34 &86.25 &89.66\\
\hline 
EfficientNetB0 &86.08 &85.00 &86.08 &90.12 \\
\rowfont{\color{blue}}
HOG + EfficientNetB0 &92.75 &90.07 &76.41 &93.13 \\
\hline 
EfficientNetB2 &87.34 &87.34&87.34 &90.24 \\
\rowfont{\color{blue}}
HOG + EfficientNetB2&92.75&94.82 &71.96 &94.49 \\

\bottomrule
\end{tabu}
\end{table*}

\begin{table*}[h!]
\centering
\caption{Accuracy results for our HOG-CNN models on ORIGA dataset with different backbones.\label{tab:origa_HOG_CNN}}
\normalsize
\setlength\tabcolsep{8 pt}
\begin{tabu}{lcccc}
\multicolumn{5}{c}{\bf{ORIGA Dataset (Glaucoma)}} \\
\toprule
\textbf{Method} & \textbf{Acc} &\textbf{Prec} & \textbf{Rec} & \textbf{AUC}  \\
\midrule

Resnet50&73.08&73.08&73.08&78.50 \\
\rowfont{\color{blue}}
HOG + Resnet50 &81.54&81.89&80.00&87.08\\
    \hline                           
DenseNet201&68.46&67.72&66.15&75.90\\
\rowfont{\color{blue}}
HOG + DenseNet201&79.23&79.07&78.46&84.06\\
    \hline
InceptionResNetV2&73.85&73.85&73.85&73.85\\
\rowfont{\color{blue}}
HOG + InceptionResNetV2&80.00&80.00&80.00&80.62\\
\hline
MobileNetV2&71.54&71.54&71.54&74.71\\
\rowfont{\color{blue}}
HOG + MobileNetV2&78.46&76.60&83.08&79.34\\
\hline
EfficientNetB3&73.85&74.05&74.62&82.46\\
\rowfont{\color{blue}}
HOG + EfficientNetB3&83.08&81.95&83.85&88.67\\
\hline
Xception&65.38&65.12&64.62&75.90\\
\rowfont{\color{blue}}
HOG + Xception&79.235&77.27&78.46&80.00\\
\hline
VGG19&69.23&68.94&70.00&76.41\\
\rowfont{\color{blue}}
HOG + VGG19&79.23&78.63&79.23&85.10\\
\hline
InceptionV3&75.38&75.59&73.85&82.88\\
\rowfont{\color{blue}}
HOG + InceptionV3&79.23&78.63&79.23&82.07\\
\hline
EfficientNetB0&70.00&70.00&70.00&77.95\\
\rowfont{\color{blue}}
HOG + EfficientNetB0&83.85&81.95&83.85&87.23\\
\hline
EfficientNetB2&64.62&65.89&65.38&67.74\\
\rowfont{\color{blue}}
HOG + EfficientNetB2&81.54&77.14&83.08&87.02\\

\bottomrule
\end{tabu}
\end{table*}

\end{appendix}

\clearpage

\bibliographystyle{elsarticle-num-names}

\bibliography{refs}

\end{document}